# Parallelizing Exploration–Exploitation Tradeoffs with Gaussian Process Bandit Optimization

**Thomas Desautels**                                              TADESAUT@CALTECH.EDU

California Institute of Technology, 1200 E California Blvd. MC 104-44, Pasadena, CA, 91125 USA

**Andreas Krause**                                                KRAUSEA@ETHZ.CH

Swiss Federal Institute of Technology Zürich, Universitätstrasse 6, 8092 Zürich, Switzerland

**Joel Burdick**                                                  JWB@ROBOTICS.CALTECH.EDU

California Institute of Technology, 1200 E California Blvd. MC 104-44, Pasadena, CA, 91125 USA

## Abstract

Can one parallelize complex exploration–exploitation tradeoffs? As an example, consider the problem of optimal high-throughput experimental design, where we wish to sequentially design batches of experiments in order to simultaneously learn a surrogate function mapping stimulus to response and identify the maximum of the function. We formalize the task as a multi-armed bandit problem, where the unknown payoff function is sampled from a Gaussian process (GP), and instead of a single arm, in each round we pull a batch of several arms in parallel. We develop GP-BUCB, a principled algorithm for choosing batches, based on the GP-UCB algorithm for sequential GP optimization. We prove a surprising result; as compared to the sequential approach, the cumulative regret of the parallel algorithm only increases by a constant factor *independent* of the batch size $B$. Our results provide rigorous theoretical support for exploiting parallelism in Bayesian global optimization. We demonstrate the effectiveness of our approach on two real-world applications.

## 1. Introduction

Many applications, from recommender systems to optimal control to experimental design, require solving exploration–exploitation tradeoffs: one needs to make a sequence of decisions with uncertain outcomes and thus, based on noisy feedback, one wishes to simultaneously learn a model and use that model to maximize the reward obtained.

Often, the set of possible decisions is large or infinite, and therefore we must be able to generalize from partial observations to predict the likely reward associated with unexplored decisions. A second crucial challenge is that we wish to explore many possible decisions in parallel: in information retrieval, it may not be possible to update the predictive model in real-time, but perhaps once per day, taking into account all the feedback collected; in experimental design, we may wish to design batches of simultaneously running experiments, only incorporating feedback once all experiments terminate; and in complex control tasks, performance feedback may become available only after a delay.

This paper tackles these two central challenges arising when solving large-scale exploration–exploitation tradeoffs. We model the problem as a stochastic multi-armed bandit problem, where the unknown mean payoff function is modeled as a Gaussian process (GP, Rasmussen & Williams (2006)). As nonparametric statistical models, GPs can flexibly incorporate a variety of assumptions about regularity of the payoff function via its covariance (or kernel) function. We design an efficient algorithm, GP-BUCB, that is able to handle both the parallel exploration problem (where we propose batches of $B$ experiments executed concurrently) and delayed feedback (where each decision can only use feedback up to $B$ rounds ago). Our approach generalizes the GP-UCB approach (Srinivas et al., 2010) to the parallel setting. We prove bounds on the cumulative regret incurred by GP-BUCB. We show that, perhaps surprisingly, near-linear speedup is possible for many commonly used kernel functions: as long as the batch size $B$ grows at most polylogarithmically in the number of rounds $T$, the GP-BUCB regret bounds only increase by a constant factor *independent of $B$* as compared to the known bounds for the sequential algorithm. We also demonstrate how the GP-BUCB algorithm can be drastically accelerated by using *lazy evaluations*. We evaluate our approach on several synthetic benchmark optimization tasks, as well as two real data sets, respectively related to automated vaccine design and therapeutic spinal cord stimulation.

**Related Work** Classical work on multi-armed bandit problems has focused on the case of a finite number of decisions (Robbins, 1952). Optimistic allocation according to upper-confidence bounds (UCB) on the payoffs has proven to be particularly effective (Auer et al., 2002). Recently, approaches for coping with large (or infinite) sets of decisions have been developed. In these cases, dependence between the payoffs associated with different decisions must be modeled and exploited. Examples include bandits with linear (Dani et al., 2008; Abernethy et al., 2008) or Lipschitz-continous payoffs (Kleinberg et al., 2008), or bandits on trees (Kocsis & Szepesvári, 2006; Bubeck et al., 2008). The exploration-exploitation tradeoff has also been studied in Bayesian global optimization and response surface modeling, where Gaussian process models are often used due to their flexibility in incorporating prior assumptions about the payoff function (Brochu et al., 2009). Several heuristics, such as Maximum Expected Improvement (Jones et al., 1998), Maximum Probability of Improvement (Mockus, 1989), and upper-confidence based methods (Cox & John, 1997), have been developed to balance exploration with exploitation and successfully applied in learning problems (Lizotte et al., 2007). Recently, Srinivas et al. (2010) analyzed GP-UCB, an upper-confidence bound sampling based algorithm for this setting, and proved bounds on its cumulative regret, and thus convergence rates for Bayesian global optimization. We build on this foundation and generalize it to the parallel setting.

To enable parallel selection, one must account for the lag between decisions and observations. Most existing approaches that can deal with such delay result in a multiplicative increase in the cumulative regret as the delay grows. Only recently, Dudik et al. (2011) demonstrated that it is possible to obtain regret bounds that only increase *additively* with the delay (i.e., the penalty becomes negligible for large numbers of decisions). However, the approach of Dudik et al. only applies to contextual bandit problems with finite decision sets, and thus not to settings with complex (even nonparametric) payoff functions. In contrast, there has been heuristic work in parallel Bayesian global optimization using GPs, e.g. by Ginsbourger et al. (2010). The state of the art is the *simulation matching* algorithm of Azimi et al. (2010). To our knowledge, no theoretical results regarding the regret of this algorithm exist. We compare with this approach in Section 5.

## 2. Problem Statement and Background

We wish to make a sequence of decisions $\boldsymbol{x}_1, \boldsymbol{x}_2, \ldots, \boldsymbol{x}_T \in D$, where $D$ is called the *decision set*, which is often (but not necessarily) a compact subset of $\mathbb{R}^d$. For each decision, we observe noisy scalar reward $y_1, y_2, \ldots, y_T$, where for any $t$, $y_t = f(\boldsymbol{x}_t) + \varepsilon_t$ and where $f : D \to \mathbb{R}$ is in turn an unknown function modeling the expected payoff $f(\boldsymbol{x})$ for each decision $\boldsymbol{x}$. For now we assume that the noise variables $\varepsilon_t$ are i.i.d. Gaussian with known variance $\sigma_n^2$, i.e., $\varepsilon_t \sim \mathcal{N}(0, \sigma_n^2)$. We will relax this assumption later. In the *strictly sequential* setting, we allow $\boldsymbol{x}_t$ to depend on observations $\boldsymbol{y}_{1:t-1}$ associated with $\boldsymbol{x}_1, \ldots, \boldsymbol{x}_{t-1}$. Below, we will formalize the main problem tackled in this paper: the challenging setting where $\boldsymbol{x}_t$ may only depend on $\boldsymbol{y}_{1:t'}$, for some $t' < t-1$.

We wish to maximize the cumulative reward $\sum_{t=1}^T f(\boldsymbol{x}_t)$, or equivalently minimize the cumulative *regret* $R_T = \sum_{t=1}^T r_t$, where $r_t = [f(\boldsymbol{x}^*) - f(\boldsymbol{x}_t)]$ and $\boldsymbol{x}^* \in \operatorname{argmax}_{\boldsymbol{x} \in D} f(\boldsymbol{x})$ is an optimum decision (assumed to exist, but not necessarily to be unique). In experimental design, $D$ might be the set of possible stimuli that can be applied, and $f(\boldsymbol{x})$ corresponds to the response to stimulus $\boldsymbol{x} \in D$. By minimizing the regret, we ensure progress towards the most effective stimulus uniformly over $T$. In fact, the average regret, $R_T/T$, is a natural upper bound on the suboptimality of the best stimulus considered so far, i.e., $R_T/T \geq \min_t [f(\boldsymbol{x}^*) - f(\boldsymbol{x}_t)]$ (often called the *simple regret*, Bubeck et al. (2009)).

**The Problem: Parallel / Delayed Selection** In many applications, we wish to select *batches* of decisions $\boldsymbol{x}_1, \ldots, \boldsymbol{x}_B$ to be evaluated in parallel. One natural application is the design of *high-throughput* experiments, where we perform several experiments in parallel, but only receive feedback after the experiments have concluded. In other settings, we may only receive feedback after a delay. In both situations, decisions are selected sequentially, but when making the decision $\boldsymbol{x}_t$ in round $t$, we can only make use of the feedback obtained in rounds $1, \ldots, t'$, for some $t' \leq t-1$. Formally, we assume there is some mapping fb $: \mathbb{N} \to \mathbb{N}_0$ such that fb$[t] \leq t-1$, $\forall t \in \mathbb{N}$, and when taking decision at time $t$, we can use feedback up to and including round fb$[t]$. If fb$[t] = 0$, no information is available.

This framework can model a variety of realistic scenarios. Setting $B = 1$ corresponds to the non-delayed, strictly sequential setting. If the feedback is delayed by exactly $B$ rounds, we can simply set $\text{fb}[t] = \max\{t - B, 0\}$. To select batches of size $B$, we can simply set $\text{fb}[t] = \lfloor (t-1)/B \rfloor B$, i.e., $\text{fb}[1] = \ldots = \text{fb}[B] = 0$, $\text{fb}[B+1] = \ldots = \text{fb}[2B] = B, \ldots$. We may also be interested in executing several experiments in parallel, but the duration of an experiment may be variable, and we can start a new experiment as soon as one finishes. In this case, $\text{fb}[t]$ may be a more complex mapping. Here, $B$ is the bound on the duration of any single experiment. In the following, we only assume that $t - \text{fb}[t] \leq B$ for some known constant $B$.

**Modeling $f$ via Gaussian Processes (GPs)** If we do not make any assumptions about the payoff function $f$, for large (possibly infinite) decision sets $D$ there is no hope to do well, i.e., incur little regret or even simply converge to an optimal decision. One effective formalism is to model $f$ as a sample from a Gaussian process (GP) prior. A GP is a probability distribution across a class of – typically smooth – functions, which is parameterized by a kernel function $k(\boldsymbol{x}, \boldsymbol{x}')$, which characterizes the smoothness of $f$, and a mean function $\mu(\boldsymbol{x})$, which we assume to be $\mu(\boldsymbol{x}) = 0$ w.l.o.g. We write $f \sim \mathcal{GP}(\mu, k)$ to denote that we model $f$ as sampled from such a GP. If we assume that the noise is i.i.d. Gaussian and we condition on a set of observations $\boldsymbol{y}_{1:t-1} = [y_1, \ldots, y_{t-1}]$ corresponding to $X = \{\boldsymbol{x}_1, \ldots, \boldsymbol{x}_{t-1}\}$, at any $\boldsymbol{x} \in D$, we obtain a Gaussian posterior $f(\boldsymbol{x})|\boldsymbol{y}_{1:t-1} \sim \mathcal{N}(\mu_{t-1}(\boldsymbol{x}), \sigma_{t-1}^2(\boldsymbol{x}))$, where

$$\mu_{t-1}(\boldsymbol{x}) = \mathbf{k}[\mathbf{K} + \sigma^2 I]^{-1} \boldsymbol{y}_{1:t-1} \text{ and} \qquad (1)$$

$$\sigma_{t-1}^2(\boldsymbol{x}) = k(\boldsymbol{x}, \boldsymbol{x}) - \mathbf{k}[\mathbf{K} + \sigma_n^2 I]^{-1} \mathbf{k}^T, \qquad (2)$$

where $\mathbf{k} = \mathbf{k}(\boldsymbol{x}, X)$ is the row vector of kernel evaluations between $\boldsymbol{x}$ and $X$ and $\mathbf{K} = \mathbf{K}(X, X)$ is the matrix of kernel evaluations between past observations.

**The GP-UCB approach** Modeling $f$ as a sample from a GP has the major advantage that the predictive uncertainty can be used to guide exploration and exploitation. Recently, Srinivas et al. (2010) analyzed the *Gaussian process Upper Confidence Bound (GP-UCB)* selection rule

$$\boldsymbol{x}_t = \operatorname*{argmax}_{\boldsymbol{x} \in D} \left[ \mu_{t-1}(\boldsymbol{x}) + \alpha_t^{1/2} \sigma_{t-1}(\boldsymbol{x}) \right]. \qquad (3)$$

This decision rule uses $\alpha_t$, a domain-specific time-varying parameter, to trade off exploitation (sampling $\boldsymbol{x}$ with high mean) and exploration (sampling $\boldsymbol{x}$ with high standard deviation) by changing the relative weighting of the posterior mean and standard deviation, respectively $\mu_{t-1}(\boldsymbol{x})$ and $\sigma_{t-1}(\boldsymbol{x})$ from Equations (1) and (2). Srinivas et al. (2010) showed that,

---

**Algorithm 1 GP-BUCB**

**Input:** Decision set $D$, GP prior $\mu_0, \sigma_0$, kernel function $k(\cdot, \cdot)$
**for** $t = 1, 2, \ldots, T$ **do**
    Choose $\boldsymbol{x}_t = \operatorname{argmax}_{\boldsymbol{x} \in D}[\mu_{\text{fb}[t]}(\boldsymbol{x}) + \beta_t^{1/2} \sigma_{t-1}(\boldsymbol{x})]$
    Compute $\sigma_t(\cdot)$
    **if** $t = \text{fb}[t+1]$ **then**
        Obtain $y_{t'} = f(\boldsymbol{x}_{t'}) + \varepsilon_{t'}$ for $t' \in \{\text{fb}[t], \ldots, t\}$
        Perform Bayesian inference to obtain $\mu_t(\cdot)$
    **end if**
**end for**

---

with proper choice of $\alpha_t$, the cumulative regret of GP-UCB grows sublinearly for many commonly used kernel functions, providing the first regret bounds and convergence rates for GP optimization.

Motivated by the strong theoretical and empirical performance of GP-UCB, we explore generalizations to batch / parallel selection (i.e., $B > 1$). One naïve approach would be to update the GP-UCB score (3) only once new feedback becomes available, but this algorithm would simply select the same observation up to $B$ times, leading to limited exploration. To encourage more exploration, one may require that no decision is selected twice (i.e., simply rank decisions according to the GP-UCB score, and pick decisions in order of decreasing score, until new feedback is available). However, since $f$ often varies smoothly, so does the GP-UCB score; this modification would also suffer from limited exploration. In the following, we introduce the Gaussian process - Batch Upper Confidence Bound (GP-BUCB) algorithm, which encourages diversity in exploration, and prove strong performance guarantees.

## 3. The GP-BUCB Algorithm

A key property of GPs is that the predictive variance (2) only depends on *where* the observations are made, but not *which* values were actually observed. Thus, it is possible to compute the posterior variance used in the sequential GP-UCB score, even while previous observations are not yet available. A natural approach towards parallel exploration is therefore to alter (3) to sequentially choose decisions within the batch as

$$\boldsymbol{x}_t = \operatorname*{argmax}_{\boldsymbol{x} \in D} \left[ \mu_{\text{fb}[t]}(\boldsymbol{x}) + \beta_t^{1/2} \sigma_{t-1}(\boldsymbol{x}) \right]. \qquad (4)$$

Here, the role of $\beta_t$ is analogous to that of $\alpha_t$ in the GP-UCB algorithm. This approach naturally encourages diversity in exploration by taking into account the change in predictive variance: since the payoffs of "similar" decisions have similar predictive distributions, exploring one decision will automatically reduce the predictive variance of similar decisions.

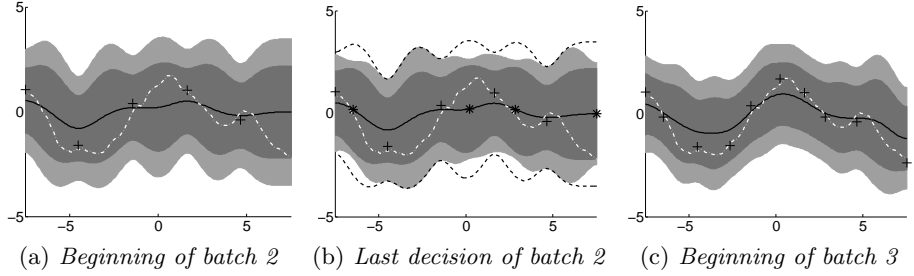

(a) *Beginning of batch 2*   (b) *Last decision of batch 2*   (c) *Beginning of batch 3*

*Figure 1.* **(a)**: The confidence intervals $C_{\mathrm{fb}[t]}^{\mathrm{seq}}(\boldsymbol{x})$ (dark), computed from previous noisy observations (crosses), are centered around the posterior mean (solid black) and contain $f(\boldsymbol{x})$ (white dashed) w.h.p. To avoid overconfidence, GP-BUCB chooses $C_{\mathrm{fb}[t]}^{\mathrm{batch}}(\boldsymbol{x})$ (light gray) such that even in the worst case, $C_t^{\mathrm{batch}}(\boldsymbol{x})$ will contain $C_{\mathrm{fb}[t]}^{\mathrm{seq}}(\boldsymbol{x})$. **(b)**: Due to the observations that GP-BUCB "hallucinates" (stars), the outer posterior confidence intervals $C_t^{\mathrm{batch}}(\boldsymbol{x})$ shrink from their values at the start of the batch (black dashed), but still contain $C_{\mathrm{fb}[t]}^{\mathrm{seq}}(\boldsymbol{x})$, as desired. **(c)**: Upon selection of the last decision of the batch, the feedback for all decisions is obtained, and new confidence intervals $C_{\mathrm{fb}[t']}^{\mathrm{seq}}(\boldsymbol{x})$ and corresponding $C_{\mathrm{fb}[t']}^{\mathrm{batch}}(\boldsymbol{x})$ are computed.

The disadvantage of taking this approach, however, is that the decision selection late in the batch is predicated on having information from the early decisions in the batch, but we do not in fact currently have that information; we are being "overconfident" about our knowledge of the function at those locations. This overconfidence requires us to compensate in a principled manner. One conceptual approach to doing so is to increase the width of the confidence intervals (through proper choice of $\beta_t$), such that the confidence intervals used by GP-BUCB are *conservative*, i.e., contain the true function $f(\boldsymbol{x})$ with high probability. Figure 1 illustrates this idea. In Section 4, we show how it is indeed possible to properly choose $\beta_t$ so that the regret only mildly increases, providing strong theoretical evidence about the potential for parallelizing GP optimization.

**Lazy Variance Calculation**  One major computational bottleneck of applying GP-BUCB is calculating the posterior mean $\mu_t(\boldsymbol{x})$ and variance $\sigma_t^2(\boldsymbol{x})$ for the candidate decisions. The mean is updated only whenever feedback is obtained, and – upon computation of the Cholesky factorization of $\mathbf{K}(X, X) + \sigma_n^2 I$ (which only needs to be done once whenever new feedback arrives) – predicting $\mu_t(\boldsymbol{x})$ takes $O(t)$ additions and multiplications. On the other hand, $\sigma_t^2$ must be recomputed for every $\boldsymbol{x}$ in $D$ after every single round, and requires solving backsubstitution, which requires $O(t^2)$ computations. Therefore, the variance computation dominates the computational cost of GP-BUCB.

Fortunately, for any fixed decision $\boldsymbol{x}$, $\sigma_t^2(\boldsymbol{x})$ is monotonically decreasing in $t$. This fact can be exploited to dramatically improve the running time of GP-BUCB, at least for finite (or when using discretizations of the) decision sets $D$. The key idea is that instead of recomputing $\sigma_{t-1}(\boldsymbol{x})$ for all decisions $\boldsymbol{x}$ in every round $t$, we can maintain an upper bound $\widehat{\sigma}_{t-1}(\boldsymbol{x})$, initial-ized to $\widehat{\sigma}_0(\boldsymbol{x}) = \infty$. In every round, we lazily apply the GP-BUCB rule with this upper bound, to identify

$$\boldsymbol{x}_t = \operatorname*{argmax}_{\boldsymbol{x} \in D}\left[\mu_{\mathrm{fb}[t]}(\boldsymbol{x}) + \beta_t^{1/2}\widehat{\sigma}_{t-1}(\boldsymbol{x})\right]. \quad (5)$$

We then recompute $\widehat{\sigma}_{t-1}(\boldsymbol{x}_t) \leftarrow \sigma_{t-1}(\boldsymbol{x}_t)$. If $\boldsymbol{x}_t$ still lies in the argmax of (5), we have identified the next decision to make, and set $\widehat{\sigma}_t(\boldsymbol{x}) = \widehat{\sigma}_{t-1}(\boldsymbol{x})$ for all remaining decisions $\boldsymbol{x}$. This idea generalizes to the bandit setting a technique proposed by Minoux (1978), which concerns calculating the greedy action for submodular maximization and leads to dramatically improved empirical computational speed, discussed in Section 5.

## 4. Regret Bounds

Srinivas et al. (2010) prove that the cumulative regret of the strictly sequential GP-UCB can be bounded (up to logarithmic factors) as $R_T = O^*(\sqrt{T\alpha_T\gamma_T})$, where

$$\gamma_T = \max_{|A| \leq T} I(f; \mathbf{y}_A) \quad (6)$$

is the maximum mutual information

$$I(f; \mathbf{y}_A) = H(\mathbf{y}_A) - H(\mathbf{y}_A \mid f) = \frac{1}{2}\log\left|\mathbf{I} + \sigma_n^{-2}\mathbf{K}(A, A)\right|$$

obtained through observations $\mathbf{y}_A$ of any set $A \subseteq D$ of $T$ decisions evaluated. For many kernel functions commonly used in practice, they show that $\gamma_T$ grows sublinearly and $\alpha_T$ only needs to grow polylogarithmically in $T$. Thus, $R_T/T \to 0$, i.e., GP-UCB is a no-regret algorithm.

The analysis of GP-UCB (and upper-confidence index policies in general) rests upon three major pillars: (1) The constructed confidence intervals

$$C_t^{\mathrm{seq}}(\boldsymbol{x}) = \left[\mu_{t-1}(\boldsymbol{x}) \pm \alpha_t^{1/2}\sigma_{t-1}(\boldsymbol{x})\right] \quad (7)$$

contain the true payoff $f(\boldsymbol{x})$ with high probability; (2) The width of the confidence interval at the selected decision bounds the instantaneous regret $r_t$ (i.e., $r_t \leq w_t$, where $w_t = 2\alpha_t^{1/2}\sigma_{t-1}(\boldsymbol{x}_t)$); and (3) the widths $w_1, \ldots, w_T$ shrink sufficiently quickly to ensure sublinear regret.

Our strategy for choosing $\beta_t$ in the GP-BUCB rule rests on a generalization of this analysis. We will choose $\beta_t$ such that the confidence intervals

$$C_t^{\text{batch}}(\boldsymbol{x}) = \left[\mu_{\text{fb}[t]}(\boldsymbol{x}) \pm \beta_t^{1/2}\sigma_{t-1}(\boldsymbol{x})\right] \quad (8)$$

still contain the true expected payoff $f(\boldsymbol{x})$ with high probability. Under this condition, a straightforward generalization of the arguments of Srinivas et al. (2010) leads to regret bounds of the form $O^*(\sqrt{T\beta_T\gamma_T})$ (formal statement is given below).

**Avoiding Overconfidence.** We seek to derive sufficient conditions on $\beta_t$ to ensure that the confidence intervals employed by GP-BUCB contain $f$ with high probability. As we will see, a crucial role is played by the *conditional mutual information*, which for observations $\boldsymbol{y}_A$ and $\boldsymbol{y}_S$ of two finite sets $A, S \subseteq D$ is defined as

$$I(f; \boldsymbol{y}_A \mid \boldsymbol{y}_S) = H(\boldsymbol{y}_A \mid \boldsymbol{y}_S) - H(\boldsymbol{y}_A \mid f).$$

Lemma 1 is the key technical result, which allows us to infer how much the confidence intervals must be enlarged to avoid overconfidence.

**Lemma 1.** *For $f$ sampled from a known GP prior with known noise variance $\sigma_n^2$, the ratio of $\sigma_{\text{fb}[t]}(\boldsymbol{x})$ to $\sigma_{t-1}(\boldsymbol{x})$ is bounded as*

$$\frac{\sigma_{\text{fb}[t]}(\boldsymbol{x})}{\sigma_{t-1}(\boldsymbol{x})} \leq \exp\{I(f; \boldsymbol{y}_{\text{fb}[t]+1:t-1} \mid \boldsymbol{y}_{1:\text{fb}[t]})\}. \quad (9)$$

Therefore, the relative amount by which the confidence intervals can shrink w.r.t. decision $\boldsymbol{x}$ is bounded by the worst-case (greatest) mutual information $I(f; \boldsymbol{y}_{\text{fb}[t]+1:t-1} \mid \boldsymbol{y}_{1:\text{fb}[t]})$ obtained during selection of $\boldsymbol{x}_{\text{fb}[t]+1:t-1}$, those decisions for which feedback is not available. Thus, if we have a constant bound $C$ on the maximum conditional mutual information that can be accrued within a batch, we can use it to guide our choice of $\beta_t$ to ensure that the algorithm is not overconfident. We can then leverage the machinery of Srinivas et al. (2010) to derive our regret bound below.

**Regret Bounds** Our main result bounds the regret of GP-BUCB in terms of a bound $C$ on the maximum conditional mutual information. It holds under any of three different assumptions about the payoff function $f$, which may all be of practical interest. In particular, it holds even if the assumption that $f$ is sampled from a GP is replaced by the assumption that $f$ has low norm in the Reproducing Kernel Hilbert Space (RKHS) associated with the kernel function.

**Theorem 1.** *Let $\delta \in (0,1)$. Suppose one of the following assumptions holds:*

1. *$D$ is finite, $f$ is sampled from a known GP prior with known noise variance $\sigma_n^2$, and $\alpha_t = 2\log(|D|t^2\pi^2/6\delta)$.*

2. *$D \subseteq [0,l]^d$ is compact and convex, $d \in \mathbb{N}$, $l > 0$. $f$ is sampled from a known GP prior with known noise variance $\sigma_n^2$, and $k(\boldsymbol{x},\boldsymbol{x}')$ satisfies the following bound w.h.p. on the derivatives of GP sample paths $f$: for some constants $a, b > 0$,*

   $$\Pr\left\{\sup_{\boldsymbol{x} \in D}|\partial f/\partial x_j| > L\right\} \leq ae^{-(L/b)^2}, \; j = 1, \ldots, d.$$

   *Choose* $\alpha_t = 2\log(t^2 2\pi^2/(3\delta)) + 2d\log\left(t^2 dbl\sqrt{\log(4da/\delta)}\right)$.

3. *$D$ is arbitrary; $f$ has RKHS norm $||f||_k \leq M$. The noise $\varepsilon_t$ form an* arbitrary *martingale difference sequence (meaning that $\mathbb{E}[\varepsilon_t \mid \varepsilon_1, \ldots, \varepsilon_{t-1}] = 0$ for all $t \in \mathbb{N}$), uniformly bounded by $\sigma_n$. Further define $\alpha_t = 2M^2 + 300\gamma_t \ln^3(t/\delta)$.*

*Further suppose we have bound $C > 0$ s.t., for all $t$,*

$$\max_{A \subseteq D, |A| \leq B-1} I(f; \boldsymbol{y}_A \mid \boldsymbol{y}_{1:\text{fb}[t]}) \leq C. \quad (10)$$

*Then, the cumulative regret of GP-BUCB, using $\beta_t = \exp(2C)\alpha_{\text{fb}[t]}$, is bounded by $O^*(\sqrt{T\gamma_T\exp(2C)\alpha_T})$ w.h.p. Precisely,*

$$\Pr\left\{R_T \leq \sqrt{C_1 T\exp(2C)\alpha_T\gamma_T} + 2 \quad \forall T \geq 1\right\} \geq 1-\delta$$

*where $C_1 = 8/\log(1 + \sigma_n^{-2})$.*

The key quantity that controls the regret in Theorem 1 is the bound $C$ on the maximum conditional mutual information obtainable within a batch (10). In particular, the cumulative regret bound of GP-BUCB is a factor $\exp(C)$ larger than the regret bound for the sequential ($B = 1$) GP-UCB algorithm. Intuitively, one expects that $C$ must grow monotonically with $B$: with greater delay, there is more potential for exploration (and thus to gain more information). An easy upper bound is obtained as follows: Due to the "information never hurts" bound (Cover & Thomas, 1991), the conditional mutual information $I(f; \boldsymbol{y}_A \mid \boldsymbol{y}_S)$ is monotonically decreasing in $S$ (i.e., as elements are added to set $S$). Therefore, $I(f; \boldsymbol{y}_A \mid \boldsymbol{y}_S) \leq I(f; \boldsymbol{y}_A) \leq \gamma_{B-1}$, whenever $|A| \leq B - 1$. However, the choice $C = \gamma_{B-1}$ is not satisfying; usually, $\gamma_{B-1}$ grows at least as $\Omega(\log B)$, suggesting that $\exp(C)$ would have to grow at least linearly in $B$. In the following, we show that it is possible to slightly modify the GP-BUCB algorithm so that a constant choice of $C$ *independent* of $B$ suffices.

*Table 1.* Initialization set sizes for Theorem 2.

| Kernel Type | Size $T^{\text{init}}$ of Initialization Set $D^{\text{init}}$ | Regret Multiplier $C'$ |
|---|---|---|
| LINEAR: $\gamma_t \leq \eta d \log(t+1)$ | $\max\left[\log(B),\ e \cdot \frac{\log \eta + \log d + 2\log(B)}{2\log(B) - 1} \eta d(B-1)\log(B)\right]$ | $\exp(2/e)$ |
| MATÉRN: $\gamma_t \leq \nu t^{\epsilon}$ | $(\nu(B-1))^{1/(1-\epsilon)}$ | $e$ |
| RBF: $\gamma_t \leq \eta(\log(t+1))^d$ | $\max\left[(\log(B))^d,\ \left(\frac{e}{d}\frac{\log \eta + (d+1)\log(B)}{2\log(B)-1}\right)^d \eta(B-1)(\log(B))^d\right]$ | $\exp((2d/e)^d)$ |

**Better Bounds Through "Initialization"** The key idea that allows us to obtain regret bounds *independent* of $B$ is again to exploit monotonicity properties of the conditional mutual information. Suppose that instead of GP-BUCB, we use a two-stage procedure, that first nonadaptively (i.e., without any feedback) selects an *initialization set* $D^{\text{init}}$ of size $|D^{\text{init}}| = T^{\text{init}}$. The algorithm then obtains feedback $\boldsymbol{y}^{\text{init}}$ for all decisions $D^{\text{init}} = \{\boldsymbol{x}_1^{\text{init}}, \ldots, \boldsymbol{x}_{T^{\text{init}}}^{\text{init}}\}$. In a second stage, it then applies GP-BUCB on the posterior Gaussian process distribution, conditioned on $\boldsymbol{y}^{\text{init}}$.

Notice that if we define

$$\gamma_T^{\text{init}} = \max_{A \subseteq D, |A| \leq T} I(f; \boldsymbol{y}_A \mid \boldsymbol{y}^{\text{init}}),$$

then, under the assumptions of Theorem 1, using $C = \gamma_{B-1}^{\text{init}}$, the regret of the two-stage algorithm is bounded by $R_T = \mathcal{O}(T^{\text{init}} + \sqrt{T \gamma_T^{\text{init}} \alpha_T \exp 2C})$. In the following, we show that it is indeed possible to construct an initialization set $D^{\text{init}}$ such that the size $T^{\text{init}}$ is dominated by $\sqrt{T \gamma_T^{\text{init}} \alpha_T \exp(2C)}$, and – crucially – that $C = \gamma_{B-1}^{\text{init}}$ can be bounded *independently* of the batch size $B$.

We will construct $D^{\text{init}}$ via uncertainty sampling: we start with $D_0^{\text{init}} = \{\}$, and for each $t = 1, \ldots, T^{\text{init}}$ greedily add the most uncertain decision

$$\boldsymbol{x}_t^{\text{init}} = \operatorname*{argmax}_{\boldsymbol{x} \in D} \sigma_{t-1}^2(\boldsymbol{x}),$$

and set $D_t^{\text{init}} = D_{t-1}^{\text{init}} \cap \boldsymbol{x}_t^{\text{init}}$. We have the following key result about the residual information gain $\gamma^{\text{init}}$:

**Lemma 2.** *Suppose we use uncertainty sampling to generate an initialization set $D^{init}$ of size $T^{init}$. Then*

$$\gamma_{B-1}^{init} \leq \frac{B-1}{T^{init}} \gamma_{T^{init}}.$$

Whenever $\gamma_T$ is sublinear (i.e., $\gamma_T = o(T)$), then for *any* constant $C > 0$, we can choose $T^{\text{init}}$ as a function of $B$ such that $\gamma_{B-1}^{\text{init}} < C$. In order to derive bounds on $T^{\text{init}}$, we in turn need a concrete analytical bound on $\gamma_T$. Fortunately, Srinivas et al. (2010) prove bounds on how the information gain $\gamma_T$ grows for some of the most commonly used kernels. Table 1 provides sufficient conditions for how quickly $T^{\text{init}}$ must grow as a function of the batch size $B$. Finally, note that uncertainty sampling is a special case of the GP-BUCB algorithm with a constant prior mean of 0 and the require-ment that for all $1 \leq t \leq T^{\text{init}}$, fb$[t] = 0$, i.e., no feedback is taken into account for the first $T^{\text{init}}$ iterations.

We summarize our analysis in the following theorem. For sake of notation, define $R_T^{\text{seq}}$ to be the regret bound of Srinivas et al. (2010) associated with the sequential GP-BUCB algorithm (i.e., Theorem 1 with $B = 1$).

**Theorem 2.** *Suppose one of the conditions of Theorem 1 is satisfied. Further suppose the kernel and $T^{init}$ are as listed in Table 1. Fix $\delta > 0$. Let $R_T$ be the regret of GP-BUCB, which ignores feedback for the first $T^{init}$ rounds. Then there exists a constant $C'$ independent of $B$ such that for any $T \geq 0$, it holds with probability at least $1 - \delta$ that*

$$R_T \leq C' R_T^{seq} + 2\|f\|_\infty T^{init},$$

*where $C'$ takes the value shown in Table 1.*

Notice that, whenever $B = O(\text{polylog}(T))$, $T^{\text{init}} = \mathcal{O}(\text{polylog}(T))$. Further note $R_T^{\text{seq}} = \Omega(\sqrt{T})$. Thus, as long as the batch size does not grow too quickly, the term $\mathcal{O}(T^{\text{init}})$ is dominated by $C' R_T^{\text{seq}}$ and thus the regret bounds of GP-BUCB are only a constant factor *independently* of $B$ worse than those of GP-UCB.

## 5. Experiments

We empirically evaluate GP-BUCB on several synthetic benchmark problems as well as two real applications. We compare it with four alternatives: (1) The strictly sequential GP-UCB algorithm ($B = 1$); (2) NRB-UCB, an approach that simply picks the maximizer of the GP-UCB score $B$ times; (3) NTB-UCB, an approach that picks the top $B$ scores according to the GP-UCB criterion; (4) A state of the art algorithm for Batch Bayesian optimization proposed by Azimi et al. (2010), which can use either a UCB or Maximum Expected Improvement (MEI) decision rule, herein SM-UCB and SM-MEI respectively. All batch selection algorithms pick batches of $B = 10$ points and all experiments were repeated for 100 trials with independent observation noise for each trial.

**Synthetic Benchmark Problems** We first test GP-BUCB in conditions where the true prior is known. A set of 100 example functions was drawn from a zero-mean GP with Matérn kernel over the interval $[0, 1]$.

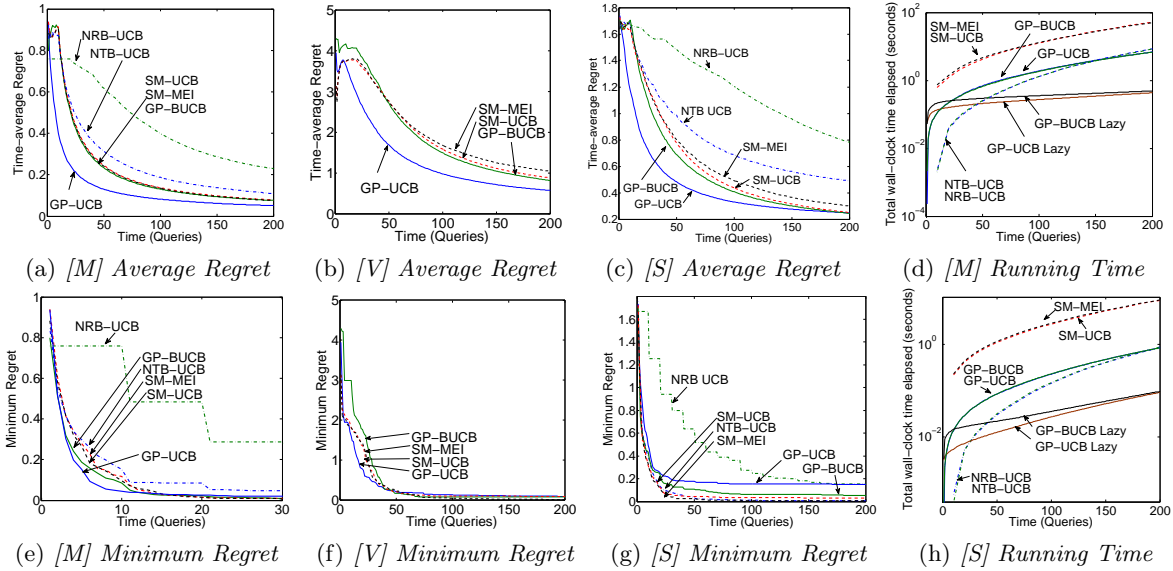

*Figure 2.* Results for [M] Matérn GP, [S] Spinal Cord Therapy and [V] Vaccine Design data sets. Average and Minimum regret are over the actions taken by the algorithm so far. All plots are averages over 100 trials. In Figures (d) and (h), note the logarithmic vertical scaling and the order of magnitude differences in run time between algorithms.

The kernel, its parameters, and the noise variance were known to each algorithm. The decision set $D$ was the discretization of $[0, 1]$ into 1000 evenly spaced points. Figures 2(a) and 2(e) present the results of this experiment. GP-BUCB performed slightly better than SM-UCB and SM-MEI in terms of both average regret and minimum regret. GP-BUCB, SM-UCB, and SM-MEI were outperformed by GP-UCB early on, but after they received their first observations at the end of batch 1 (query 10), performance was comparable to GP-UCB. As expected, both of the naïve algorithms performed quite poorly. Figure 2(d) compares the algorithms in terms of their running time; lazy variance calculations led to dramatic running time improvements. We also performed experiments on other synthetic benchmark domains, with qualitatively similar results (presented in the supplemental material).

**Automated Vaccine Design** We also tested GP-BUCB on a database of Widmer et al. (2010), which describes the binding affinity of various peptides with a Major Histocompatibility Complex (MHC) Class I molecule, of importance when designing vaccines to exploit peptide binding properties. Each of the peptides is described by a set of chemical features in $\mathbb{R}^{45}$. The binding affinity of each peptide, which is treated as the reward or payoff, is described as an offset $IC_{50}$ value. The experiments used a linear ARD kernel fitted on a different MHC molecule from the same data set. Figures 2(b) and 2(f) present this experiment's results. GP-BUCB performs competitively with SM-MEI and SM-UCB, both in terms of average and minimum regret, and converges to the performance of GP-UCB.

**Spinal Cord Therapy** Lastly, we compare the algorithms on a data set of leg muscle activity triggered by therapeutic spinal electrostimulation in spinal cord injured rats. The experimental objective is to choose the stimulus electrodes which maximize the resulting activity in lower limb muscles, as measured by electromyography (EMG), in order to improve spinal reflex and locomotor function. We sought to maximize the peak-to-peak amplitude of the recorded EMG waveforms from the right medial Gastrocnemius muscle in a time window corresponding to a single interneuronal delay. This objective function measures to what degree the selected stimulus activates the interneurons in the spinal gray matter which control reflex activity. Electrode configurations were represented in $\mathbb{R}^4$ by the cathode and anode locations on the array. A squared-exponential ARD kernel was fitted for this space using experimental data from 12 days post-injury. Algorithm testing was done on data from 116 electrode pairs tested on the 14th day post-injury. Experimental results are presented in Figures 2(c) and 2(g). This problem setting was quite challenging for all algorithms, as the data was highly multi-modal. Consequently, GP-UCB often failed to find the optimum in the number of queries examined; out of 100 runs, only 18 had converged to the optimum, and out of the remainder, none had ever visited the optimum in 200 queries. Interestingly, the GP-BUCB, SM-UCB and SM-MEI algorithms were more robust to these difficulties; their superior initialization, born of the exploratory behavior forced on them by their initial ignorance, resulted in convergence likelihoods on the or-der of 40% for each. The superior performance of GP-BUCB to SM-UCB and SM-MEI with respect to average regret and the comparable likelihoods of convergence within the practical experimental window considered indicate that GP-BUCB is at least as effective as the current state of the art in this challenging experimental setting. Lazy variance calculations again led to dramatic running time improvements, presented in Figure 2(h).

## 6. Conclusions

We have developed the GP-BUCB algorithm for parallelizing exploration and exploitation tradeoffs in Gaussian process bandit optimization. We showed how the regret of GP-BUCB can be bounded in terms of an intuitive conditional mutual information quantity. Using this analysis, we prove that GP-BUCB can be "initialized" to obtain regret bounds which only additively depend on the batch size for many kernel functions commonly used. We further show how "lazy" variance evaluation can yield order-of-magnitude improvements in running time. In our experiments, GP-BUCB compares favorably to the state of the art in parallel Bayesian optimization, which is not equipped with theoretical guarantees. We believe that our results provide an important step towards solving complex, large-scale exploration-exploitation tradeoffs.

**Acknowledgments** The authors thank Daniel Golovin for helpful discussions. This work was partially supported by NIH project R01 NS062009, SNSF grant 200021_137971, NSF IIS-0953413, DARPA MSEE FA8650-11-1-7156 and the ThinkSwiss Research Scholarship.

## References

Abernethy, J., Hazan, E., and Rakhlin, A. Competing in the dark: An efficient algorithm for bandit linear optimization. In *COLT*, 2008.

Auer, P., Cesa-Bianchi, N., and Fischer, P. Finite-time analysis of the multiarmed bandit problem. *Mach. Learn.*, 47(2-3):235–256, 2002.

Azimi, J., Fern, A., and X.Fern. Batch bayesian optimization via simulation matching. In *NIPS*, 2010.

Brochu, E., Cora, M., and de Freitas, N. A tutorial on Bayesian optimization of expensive cost functions, with application to active user modeling and hierarchical reinforcement learning. In *TR-2009-23, UBC*, 2009.

Bubeck, S., Munos, R., Stoltz, G., and Szepesvári, C. Online optimization in X-armed bandits. In *NIPS*, 2008.

Bubeck, S., Munos, R., and Stoltz, G. Pure exploration in multi-armed bandits problems. In *ALT*, 2009.

Cover, T. M. and Thomas, J. A. *Elements of Information Theory*. Wiley Interscience, 1991.

Cox, D. D. and John, S. Sdo: A statistical method for global optimization. *Multidisciplinary Design Optimization: State of the Art*, 1997.

Dani, V., Hayes, T. P., and Kakade, S. M. Stochastic linear optimization under bandit feedback. In *COLT*, 2008.

Dudik, M., Hsu, D., Kale, S., Karampatziakis, N., Langford, J., Reyzin, L., and Zhang, T. Efficient optimal learning for contextual bandits. In *UAI*, 2011.

Ginsbourger, D., Riche, R., and Carraro, L. Kriging is well-suited to parallelize optimization. In Tenne, Yoel and Goh, Chi-Keong (eds.), *Computational Intelligence in Expensive Optimization Problems*, volume 2 of *Adaptation, Learning, and Optimization*, pp. 131–162. Springer Berlin Heidelberg, 2010.

Jones, D. R., Schonlau, M., and Welch, W. J. Efficient global optimization of expensive black-box functions. *J Glob. Opti.*, 13:455–492, 1998.

Kleinberg, R., Slivkins, A., and Upfal, E. Multi-armed bandits in metric spaces. In *STOC*, pp. 681–690, 2008.

Kocsis, L. and Szepesvári, C. Bandit based monte-carlo planning. In *ECML*, 2006.

Lizotte, D., Wang, T., Bowling, M., and Schuurmans, D. Automatic gait optimization with Gaussian process regression. In *IJCAI*, pp. 944–949, 2007.

Minoux, M. Accelerated greedy algorithms for maximizing submodular set functions. *Optimization Techniques, LNCS*, pp. 234–243, 1978.

Mockus, J. *Bayesian Approach to Global Optimization*. Kluwer Academic Publishers, 1989.

Rasmussen, C. E. and Williams, C. K. I. *Gaussian Processes for Machine Learning*. MIT Press, 2006.

Robbins, H. Some aspects of the sequential design of experiments. *Bul. Am. Math. Soc.*, 55, 1952.

Srinivas, N., Krause, A., Kakade, S., and Seeger, M. Gaussian process optimization in the bandit setting: No regret and experimental design. In *ICML*, 2010.

Widmer, C., Toussaint, N., Altun, Y., and Rätsch, G. Inferring latent task structure for multitask learning by multiple kernel learning. *BMC Bioinformatics*, 11(Suppl 8:S5), 2010.